\documentclass[10pt,conference,a4paper]{IEEEtran}
\ifCLASSINFOpdf
\else
\fi
\usepackage{times}
\usepackage{epsfig}
\usepackage{graphicx}
\usepackage{amsmath}
\usepackage{amssymb}
\usepackage{multirow}
\usepackage{CJK}

\usepackage{subfig}
\usepackage{color}
\usepackage[colorlinks,linkcolor=red]{hyperref}
\begin{document}
%
\title{Large-scale Continuous Gesture Recognition Using \\Convolutional Neural Networks}


\author{Pichao Wang, Wanqing Li, Song Liu, Yuyao Zhang, Zhimin Gao and Philip Ogunbona\\
\\
Advanced Multimedia Research Lab, University of Wollongong, Australia\\
\\
{ pw212@uowmail.edu.au, \{wanqing, songl\}@uow.edu.au,}\\ 
{ \{yz606, zg126@uowmail.edu.au\}, philipo@uow.edu.au}
}


%


\maketitle

\begin{abstract}

This paper addresses the problem of continuous gesture recognition from sequences of depth maps using Convolutional Neural networks (ConvNets). The proposed method first segments individual gestures from a depth sequence based on quantity of movement (QOM). For each segmented gesture, an Improved Depth Motion Map (IDMM), which converts the depth sequence into one image, is constructed and fed to a ConvNet for recognition. The IDMM effectively encodes both spatial and temporal information and allows the fine-tuning with existing ConvNet models for classification without introducing millions of parameters to learn. The proposed method is evaluated on the Large-scale Continuous Gesture Recognition of the ChaLearn Looking at People (LAP) challenge 2016. It achieved the performance of 0.2655 (Mean Jaccard Index) and ranked $3^{rd}$ place in this challenge.

\end{abstract}


%
\IEEEpeerreviewmaketitle

\section{Introduction}
Gesture and human action recognition from visual information is an active research topic in Computer Vision and Machine Learning. It has many potential applications including human-computer interactions and robotics. Since first work~\cite{li2010action} on action recognition from depth data captured by commodity depth sensors (e.g., Kinect) in 2010, many methods~\cite{wang2012mining,Yang2012a,Oreifej2013,yangsuper,
pichao2014,lurange,Vemulapallia2016,zhang2016rgba} for action recognition have been proposed based on specific hand-crafted feature descriptors extracted from depth or skeleton data. With the recent development of deep learning, a few methods have also been developed based on Convolutional Neural Networks (ConvNets)~\cite{pichao2015,pichaoTHMS,pichao2016,pichaoicprb,pichaocsvt2016} and Recurrent Neural Networks (RNNs)~\cite{du2015hierarchical,veeriah2015differential,zhu2015co,shahroudy2016ntu}.  However, most of the works on gesture/action recognition reported to date focus on the classification of individual gestures or actions by assuming that instances of individual gestures and actions have been isolated or segmented from a video or a stream of depth maps/skeletons before the classification. In the cases of continuous recognition, the input stream usually contains unknown numbers, unknown orders and unknown boundaries of gestures or actions and both segmentation and recognition have to be solved at the same time. 


There are three common approaches to continuous recognition. The first approach is to tackle the segmentation and classification of the gestures or actions separately and sequentially. The key advantages of this approach are that different features can be used for segmentation and classification and existing classification methods can be leveraged. The disadvantages are that both segmentation and classification could be the bottleneck of the systems and they can not be optimized together.  The second approach is to apply classification to a sliding temporal window and aggregate the window based classification to achieve final segmentation and classification. One of the key challenges in this approach is the difficulty of setting the size of sliding window because durations of different gestures or actions can vary significantly.  The third approach is to perform the segmentation and classification simultaneously. 

\begin{figure*}[t]
\begin{center}
{\includegraphics[height = 60mm, width = 180mm]{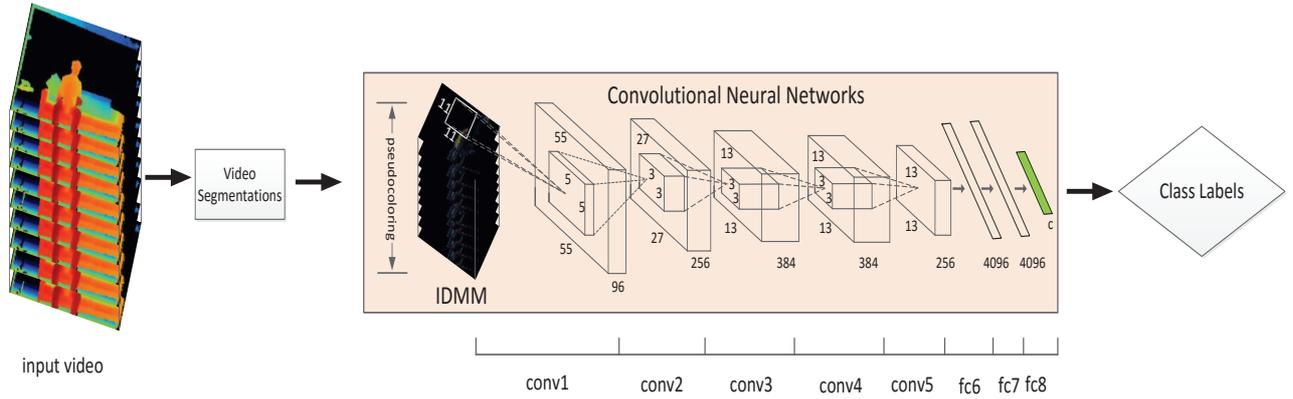}}
\end{center}
\caption{The framework for proposed method.}
\label{fig:framework}
\end{figure*}

This paper adopts the first approach and focuses on robust classification using ConvNets that are insensitive to inaccurate segmentation of gestures. Specifically,  individual gestures are first segmented based on quantity of movement (QOM)~\cite{jiang2015multi} from a stream of depth maps. For each segmented gesture, an Improved Depth Motion Map (IDMM) is constructed from its sequence of depth maps. ConvNets are used to learn the dynamic features from the IDMM for effective classification. Fig.~\ref{fig:framework} shows the framework of the proposed method.


The rest of this paper is organized as follows. Section II briefly reviews the related works on video segmentation and gesture/action recognition based on depth and deep learning. Details of the proposed method are presented in Section III. Experimental results on the dataset provided by the challenge are reported in Section IV, and Section V concludes the paper.

\section{Related Work}
\subsection{Video Segmentation}
There are many methods proposed for segmenting individual gestures from a 
video. The popular and widely used method employs dynamic time 
warping~(DTW) to decide the delimitation frames of individual 
gestures~\cite{jmlr14jun, tip14-jun, paa15-jun}.  Difference images are first 
obtained by subtracting two consecutive grey-scale images and each difference 
image is partitioned into a grid of $3\times3$ cells. Each cell is then 
represented by the average value of pixels within this cell. The matrix of the 
cells in a difference image is flattened as a vector called motion feature and 
calculated for each frame in the video excluding the final frame. This
results in a $9\times{}(K-1)$ matrix of motion features for a video with $K$ 
frames. The motion feature matrix is extracted from both test video and training 
video which consists of multiple gestures. The two matrices are treated as two 
temporal sequences with each motion feature as a feature vector at an instant 
of time. The distance between two feature vectors is defined as the negative 
Euclidean distance and a matrix containing DTW  distances (measuring similarity between two temporal sequences) between the two 
sequences is then calculated and analysed by Viterbi 
algorithm~\cite{pieee73forney} to segment the gestures. 

Another category of gesture segmentation methods from a multi-gesture video is 
based on appearance. Upon the general assumption that the \textit{start} and 
\textit{end} frames of adjacent gestures are similar, correlation 
coefficients~\cite{jmlr12-yuiman} and K-nearest neighbour algorithm with 
histogram of oriented gradient~(HOG)~\cite{cvpr12-wu} are used to identify the 
\textit{start} and \textit{end} frames of gestures.  Jiang et 
al.~\cite{jiang2015multi} proposed a method based on quantity of 
movement~(QOM) by assuming the same \textit{start} pose among different 
gestures. Candidate delimitation frames are chosen based on the global QOM. 
After a refining stage which employs a sliding window to keep the frame with 
minimum QOM in each windowing session, the \textit{start} and \textit{end} 
frames are assumed to be the remained frames. This paper adopts the QOM based 
method and its details will be presented in Section~\ref{subsec:video-seg}.

\subsection{Depth Based Action Recognition}
With Microsoft Kinect Sensors researchers have developed methods for depth map-based action recognition. Li et al. \cite{li2010action} sampled points 
from a depth map to obtain a bag of 3D points to encode spatial information and 
employ an expandable graphical model to encode temporal information 
\cite{li2008}. Yang et al. \cite{Yang2012a} stacked differences between projected depth maps as 
a depth motion map (DMM) and then used HOG to extract relevant features from the 
DMM. This method transforms the problem of action recognition from 
spatio-temporal space to spatial space.  In \cite{Oreifej2013}, a feature called Histogram of Oriented 4D 
Normals (HON4D) was proposed; surface normal is extended to 4D space and 
quantized by regular polychorons. Following this method, Yang and Tian 
\cite{yangsuper} cluster hypersurface normals and form the polynormal which can 
be used to jointly capture the local motion and geometry information. Super 
Normal Vector (SNV) is generated by aggregating the low-level polynormals. In 
\cite{lurange}, a fast binary range-sample feature was proposed based on a test 
statistic by carefully designing the sampling scheme to exclude most pixels that 
fall into the background and to incorporate spatio-temporal cues. 
\subsection{Deep Leaning Based Recogntiion}

Existing deep learning approach can be generally divided into four categories 
based on how the video is represented and fed to a deep neural network. The 
first category views a video either as a set of still 
images~\cite{yue2015beyond} or as a short and smooth transition between similar 
frames~\cite{simonyan2014two}, each color channel of the images is fed to one 
channel of a ConvNet. Although obviously suboptimal, considering the video as a 
bag of static frames performs 
reasonably well. The second category represents a video as a volume and 
extends ConvNets to a third, temporal dimension~\cite{ji20133d,tran2015learning} 
replacing 2D filters with 3D equivalents. So far, this approach has produced 
little benefits, probably due to the lack of annotated training data. The third 
category treats a video as a sequence of images and feeds the sequence to 
an 
RNN~\cite{donahue2015long,du2015hierarchical,veeriah2015differential,zhu2015co,
shahroudy2016ntu}. An RNN is typically considered as memory cells, which are 
sensitive to both short as well as long term patterns. It parses the video 
frames sequentially and encodes the frame-level information in their memory. 
However, using RNNs did not give an improvement over temporal pooling of 
convolutional features~\cite{yue2015beyond} or over hand-crafted features. The 
last category represents a video as one or multiple compact images and 
adopts available trained ConvNet architectures for 
fine-tuning~\cite{pichao2015,pichaoTHMS,pichao2016,bilen2016dynamic}. This 
category has achieved state-of-the-art results of action recognition on many RGB 
and depth/skeleton datasets. The proposed gesture classification in this paper 
falls into the last category.

\section{Proposed Method}
The proposed method consists of two major components, as illustrated in 
Fig.~\ref{fig:framework}:
video segmentation and construction of Improved Depth Motion Map (IDMM) from a 
sequence of depth maps as the input to ConvNets. Given a sequence of depth maps 
consisting of multiple gestures, the \textit{start} and \textit{end} frames 
of each gesture are identified based on quantity of movement 
(QOM)~\cite{jiang2015multi}. Then, 
one IDMM is constructed by accumulating the absolute depth difference between 
current frame and the \textit{start} frame for each gesture segment. The IDMM 
goes through a pseudo-color coding process to become a pseudo-RGB image as an 
input to a ConvNet for classification. The main objective of pseudo-color coding 
is to enchance the motin pattern captured by the IDMM.  In the rest of this 
section, video segmentation, construction of IDMM, pseudo-color coding of 
IDMM, and training of the ConvNets are explained in detail.

\begin{table*}[!ht]
\centering
\caption{Information of the ChaLearn LAP ConGD Dataset. \label{table1}}
\begin{tabular}{|c|c|c|c|c|c|c|c|}
\hline
Sets &\# of labels &\# of gestures & \# of RGB videos & \# of depth videos & \# of subjects & label provided & temporal segment provided\\
\hline
Training & 249 & 30442 & 14134 & 14134 & 17 & Yes & Yes\\
\hline
Validation & 249 & 8889 & 4179 & 4179 & 2 & No & No\\
\hline
Testing & 249 &  8602 & 4042 & 4042 & 2 & No & No\\
\hline
All & 249 & 47933 & 22535 & 22535 & 21 & - & -\\
\hline
\end{tabular}
\end{table*}

\subsection{Video Segmentation}\label{subsec:video-seg}

Given a sequence of depth maps that contains multiple gestures, 
The \textit{start} and \textit{end} frames of each gesture is detected based on 
quantity of movement~(QOM)~\cite{jiang2015multi} by assuming that all gestures 
starts from a similar pose, referred to as Neural pose. QOM between two frames 
is a binary image obtained by applying $f(\cdot)$ pixel-wisely on the 
difference image of two depth maps. $f(\cdot)$ is a step function from 0 to 1 at 
the ad hoc threshold of 60. 
The  global QOM of a frame at time $t$ is defined as QOM between Frame $t$ and 
the very first frame of the whole video sequence. A set of frame indices of 
candidate delimitation frames was initialised by choosing frames with lower 
global QOMs than a threshold. The threshold was calculated by adding the mean to 
twice the standard deviation of global QOMs extracted from first and last 
$12.5\%$ of the averaged gesture sequence length $L$ which was calculated from 
the training gestures. A sliding window with a size of $\frac{L}{2}$ was then 
used to refine the candidate set and in each windowing session only the index of 
frame with a minimum global QOM is retained. After the refinement, the remaining 
frames are expected to be the deliminating frames of gestures.

\subsection{Construction Of IDMM}

Unlike the Depth Motion Map (DMM) \cite{Yang2012a} which is calculated by
accumulating the thresholded difference between consecutive frames, two 
extensions are proposed to construct an IDMM. First, the motion energy is 
calculated by accumulating the absolute difference between the current frame and 
the Neural pose frame. This would better measure the slow motion than original 
DMM. Second, to preserve subtle motion information, the motion energy is 
calculated without thresholding. Calculation of IDMM can be expressed as:

\begin{align}
IDMM = \sum_{i=1}^{N}|(\text{depth of frame})^{i} - 
\text{depth of Neural frame}|
\end{align}
where $i$ denotes the index of the frame and $N$ represents the total number 
of frames in the segmented gesture. For simplicity, the first frame of each 
segment is considered as the Neural frame.

\subsection{Pseudocoloring}
In their work Abidi et al.~\cite{abidi2006improving} used color-coding 
to harness the perceptual capabilities of the human visual system and 
extracted more information from gray images. The detailed texture patterns in 
the image are significantly enhanced. Using this as a motivation, it is proposed 
in this paper to code an IDMM into a pseudo-color image and effectively 
exploit/enhance the texture in the IDMM that corresponds to the motion patterns 
of actions. In this work, a power rainbow transform is adopted. For a given 
intensity $I$, the power 
rainbow transform encodes it into a normalized color $(R^{*},G^{*},B^{*})$ as 
follows:
\begin{equation}\label{eq7}
  \begin{split}
    &R^{*} = \{(1 + cos(\frac{4\pi}{3*255}I))/2\}^{2}\\
    &G^{*} = \{(1 + cos(\frac{4\pi}{3*255}I-\frac{2\pi}{3}))/2\}^{2}\\
    &B^{*} = \{(1 + cos(\frac{4\pi}{3*255}I-\frac{4\pi}{3}))/2\}^{2},
  \end{split}
\end{equation}

where $R^{*}$, $G^{*}$ and $B^{*}$ are the normalized RGB values through the 
power rainbow transform. To code an IDMM, linear mapping is used to convert IDMM 
values to $I\in[0,255]$ across per image.

\begin{figure*}[t]
\begin{center}
{\includegraphics[height = 70mm, width = 180mm]{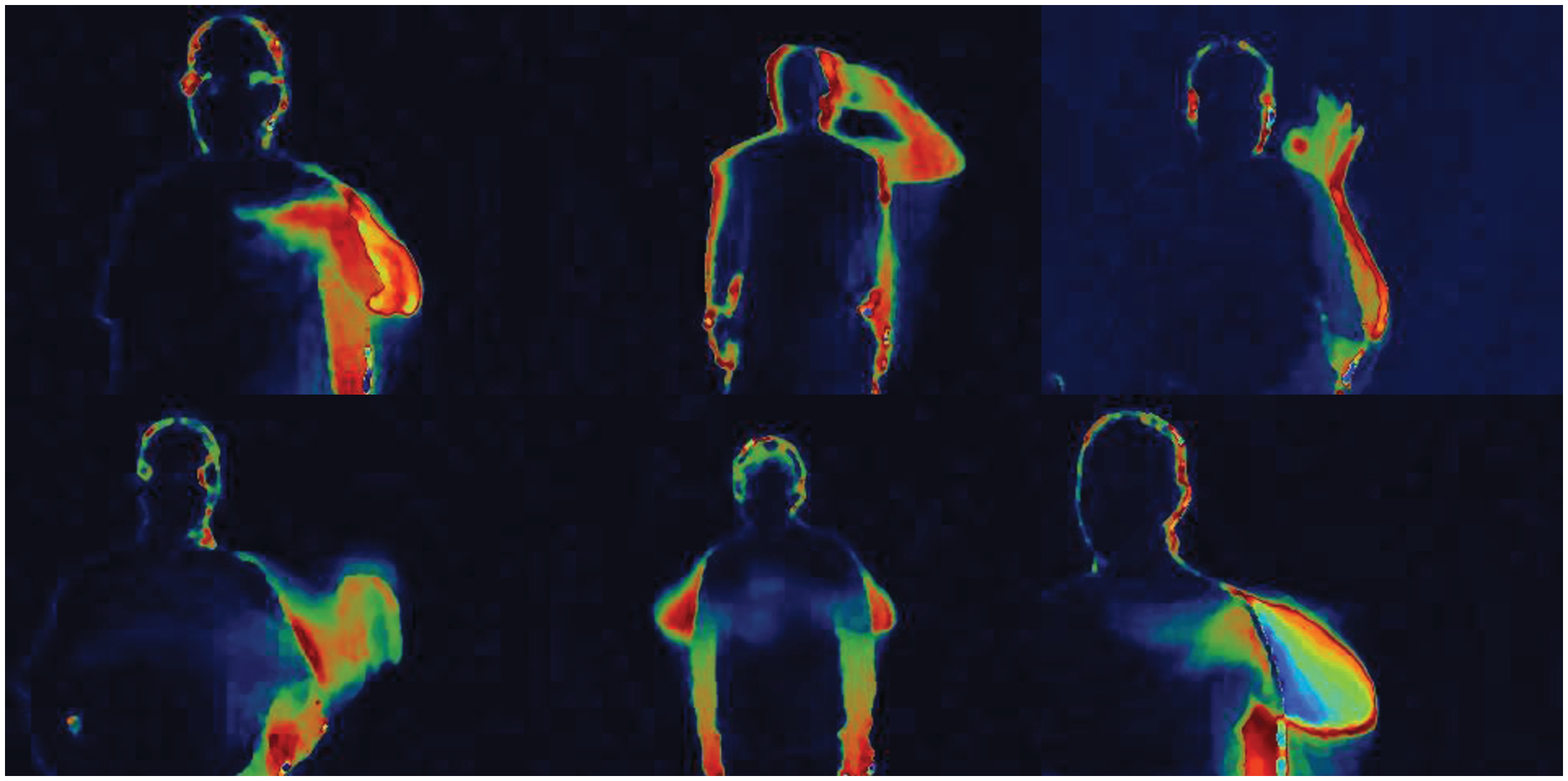}}
\end{center}
\caption{The samples of IDMMs for six different gestures. 
The labels from top left to bottom right 
are:Mudra1/Ardhachandra;Mudra1/Ardhapataka;Mudra1/Chandrakala;Mudra1/Chatura;Mud
ra1/Kartarimukha;Mudra1/Pataka.}
\label{fig:IDMM}
\end{figure*}

The resulting IDMMs are illustrated as in Fig.~\ref{fig:IDMM}.

\subsection{Network Training \& Classification}

One ConvNet is trained on the pseudo-color coded IDMM. 
The layer configuration of the ConvNets follows that 
in~\cite{krizhevsky2012imagenet}. The ConvNet contains eight layers, the first 
five are convolutional layers and the remaining three are fully-connected 
layers. The implementation is derived from the publicly available Caffe toolbox 
\cite{jia2014caffe} based on one {NVIDIA Tesla K40 GPU} card.

The training procedure is similar to that in~\cite{krizhevsky2012imagenet}. 
The network weights are learned using the mini-batch stochastic gradient descent 
with the momentum set to 0.9 and weight decay set to 0.0005. All 
hidden weight layers use the rectification (RELU) activation function. At each 
iteration, a mini-batch of 256 samples is constructed by sampling 256 shuffled 
training color-coded IDMMs. All color-coded IDMMs are resized to $256 \times 
256$. The learning rate for fine-tuning is set to $10^{-3}$ with pre-trained 
models on ILSVRC-2012, and then it is decreased according to a fixed schedule, 
which is kept the same for all training sets.  For the ConvNet the training 
undergoes 20K iterations and the learning rate decreases every 5K iterations. 
For all experiments, the dropout regularisation ratio was set to 0.5 in order to 
reduce complex co-adaptations of neurons in nets. 

Given a test depth sequence, a pseudo-colored IDMM is 
constructed for each segmented gesture and the trained ConvNet is used to 
predict the gesture label of the segment.

\section{Experiments}
In this section, the Large-scale Continuous Gesture Recognition 
Dataset of the ChaLearn LAP challenge 2016 (ChaLearn LAP ConGD Dataset))~\cite{ICPRW2016} and 
evaluation protocol are described. The experimental results of the proposed 
method on this dataset are reported and compared with the baselines recommended 
by the chellenge organisers. 

\subsection{Dataset}
\begin{figure*}[t]
\begin{center}
{\includegraphics[height = 80mm, width = 180mm]{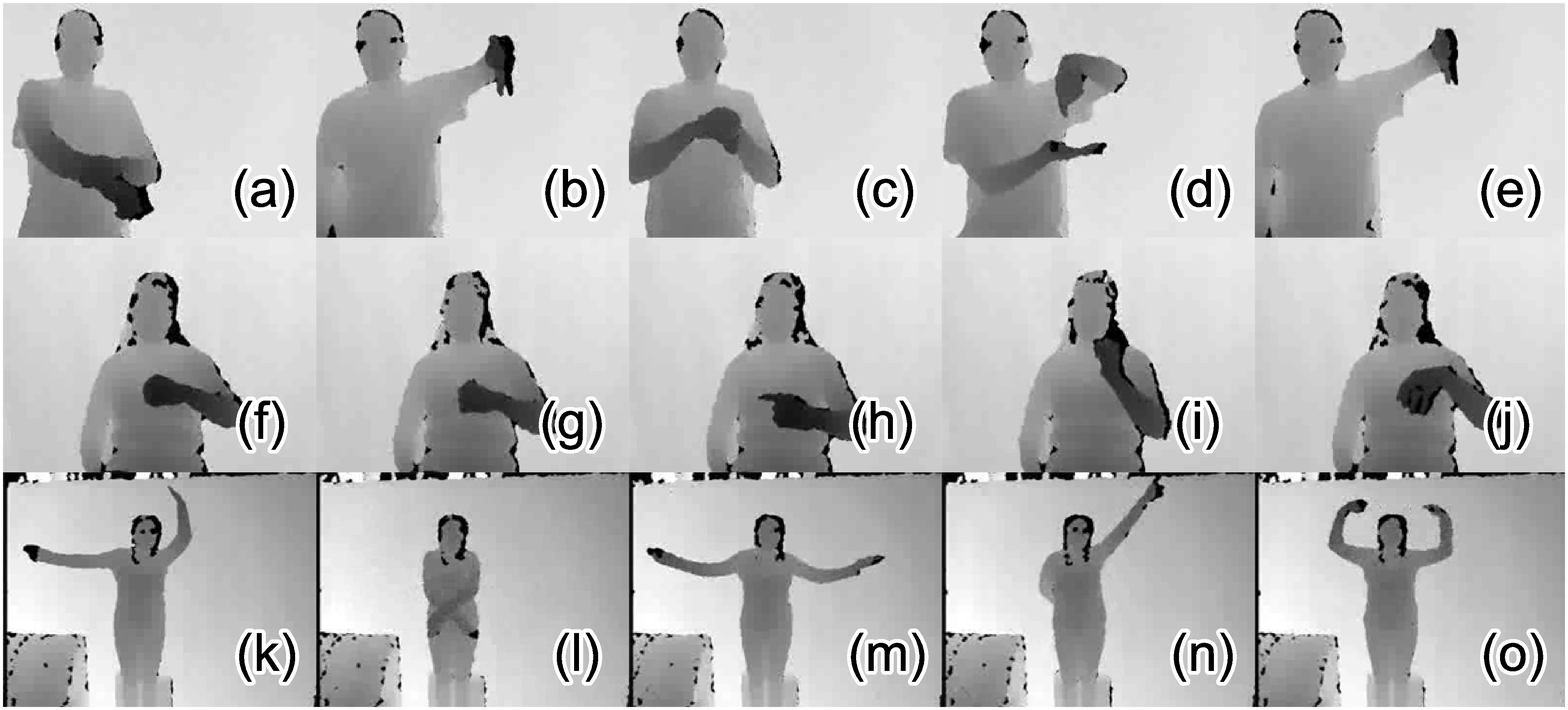}}
\end{center}
\caption{The sample depth maps from three sequences, each containing 5 different gestures. Each row corresponds to one depth video sequence. The labels from top left to bottom right are: \\(a) CraneHandSignals/EverythingSlow;(b) RefereeVolleyballSignals2/BallServedIntoNetPlayerTouchingNet;\\(c) GestunoDisaster/110\_earthquake\_tremblementdeterre;(d) DivingSignals2/You;\\(e) RefereeVolleyballSignals2/BallServedIntoNetPlayerTouchingNet;(f) Mudra2/Sandamsha;\\(g) Mudra2/Sandamsha;(h) DivingSignals2/CannotOpenReserve;(i) GestunoTopography/95\_region\_region;\\(j) DivingSignals2/Meet;(k) RefereeVolleyballSignals1/Timeout;(l) SwatHandSignals1/DogNeeded;\\(m) DivingSignals2/ReserveOpened;(n) DivingSignals1/ComeHere;\\(o) DivingSignals1/Watch, SwatHandSignals2/LookSearch.}
\label{fig:samples}
\end{figure*}

The ChaLearn LAP ConGD Dataset is derived from the ChaLearn Gesture Dataset 
(CGD)~\cite{guyon2014chalearn}. 
It has 47933 RGB-D  gesture instances in 22535 RGB-D gesture videos. Each RGB-D 
video may contain one or more gestures.  There are 249 gestures performed by 21 
different individuals.  The detailed information of this dataset is shown in 
Table~\ref{table1}. In this paper, only depth data was used in the proposed 
method. Some samples of depth maps are shown in Fig.~\ref{fig:samples}.

\subsection{Evaluation Protocol}

The dataset was divided into training, validation 
and test sets by the challenge organizers. All three sets include data from 
different subjects and the gestures of one subject in validation and test sets 
do not appear in the training set. 

Jaccard index (the higher the better) is adopted to measure the performance. 
The Jaccard index measures the average relative overlap between true and 
predicted sequences of frames for a given gesture. For a sequence $s$, let 
$G_{s,i}$ and $P_{s,i}$ be binary indicator vectors for which 1-values 
correspond to frames in which the $i^{th}$ gesture label is being performed. The 
Jaccard Index for the $i^{th}$ class is defined for the sequence $s$ as:

\begin{equation}
J_{s,i} = \dfrac{G_{s,i}\bigcap P_{s,i}}{G_{s,i}\bigcup P_{s,i}},
\end{equation}
where $G_{s,i}$ is the ground truth of the $i^{th}$ gesture label in 
sequence $s$, and $P_{s,i}$ is the prediction for the $i^{th}$ label in sequence 
$s$.

When $G_{s,i}$ and $P_{s,i}$ are empty, $J_{(s,i)}$ is defined to be 0. 
Then for the sequence $s$ with $l_{s}$ true labels, the Jaccard Index $J_{s}$ is 
calculated as:

\begin{equation}
J_{s} = \dfrac{1}{l_{s}}\sum_{i=1}^{L}J_{s,i}.
\end{equation}
For all testing sequences $S = {s_{1},...,s_{n}}$ with $n$ gestures, the mean 
Jaccard Index $\overline{J_{S}}$ is used as the evaluation criteria and 
calculated as:
 \begin{equation}
 \overline{J_{S}} = \dfrac{1}{n}\sum_{j=1}^{n}J_{s_{j}}.
 \end{equation}

\subsection{Experimental Results}

The results of the proposed method on the validation and test sets and their 
comparisons to the results of the baseline methods~\cite{pami16Jun} (MFSK and 
MFSK+DeepID) are shown in Table~\ref{table2}. The codes and models can be downloaded at the author's homepage \url{https://sites.google.com/site/pichaossites/}\\
\href{https://sites.google.com/site/pichaossites/}.

\begin{table}[!ht]
\centering
\caption{Accuracies of the proposed method and baseline 
methods on the ChaLearn LAP ConGD Dataset. \label{table2}}
\begin{tabular}{|c|c|c|}
\hline
Method & Set & Mean Jaccard Index $\overline{J_{S}}$\\
\hline
MFSK & Validation & 0.0918\\
\hline
MFSK+DeepID & Validation & 0.0902\\
\hline
Proposed Method & Validation & \textbf{0.2403}\\
\hline
MFSK & Testing & 0.1464 \\
\hline
MFSK+DeepID & Testing & 0.1435\\
\hline
Proposed & Testing &\textbf{0.2655}\\
\hline
\end{tabular}
\end{table}

The results showed that the proposed method 
significantly outperformed the baseline methods, even though only single 
modality, i.e. depth data, was used while the baseline method used both RGB and 
depth videos.  

The first three winners' results are summarized in Table~\ref{table3}. We can see that our method is among the top performers and our recognition rate is very close to the best performance of this challenge (0.265506 vs. 0.269235\&0.286915), even though we only used depth data for proposed method. Regarding computational cost, our implementation is based on CUDA 7.5 and Matlab 2015b, and it takes about 0.8s to process one depth sequence for testing in our workstation equipped with 8 cores CPU, 64G RAM, and Tesla K40 GPU.

\begin{table}[!ht]
\centering
\caption{Comparsion the performances of the first three winners in this challenge.  Our team ranks the third place in the ICPR ChaLearn LAP challenge 2016. \label{table3}}
\begin{tabular}{|c|c|c|}
\hline
Rank & Team & Mean Jaccard Index $\overline{J_{S}}$\\
\hline
1 & ICT\_NHCI~\cite{xiujuan} & 0.286915\\
\hline
2 & TARDIS~\cite{Necati} & 0.269235\\
\hline
3 & \textbf{AMRL (ours)} & 0.265506\\
\hline
\end{tabular}
\end{table}

\section{Conclusions}

This paper presents an effective yet simple method 
for continuous gesture recognition using only depth map sequences. Depth 
sequences are first segmented so that each segmentation contains only one 
gesture and a ConvNet is used for feature extraction and classification. The 
proposed construction of IDMM enables the use of available pre-trained models 
for fine-tuning without learning afresh. Experimental results on 
ChaLearn LAP ConGD Dataset verified the effectiveness of the proposed method. How to exactly extract the neutral pose and fuse different modalities to improve the accuracy will be our future work.

\section*{Acknowledgment}

The authors would like to thank NVIDIA Corporation for the donation of a Tesla K40 GPU card used in this challenge.

\end{document}